%% file: main.tex
\documentclass{article}

\usepackage[margin=1in]{geometry}

\usepackage[T1]{fontenc}
\usepackage[utf8]{inputenc}

\usepackage{microtype}
\usepackage{booktabs}
\usepackage{multirow}
\usepackage{amsmath,amssymb}
\usepackage{graphicx}
\usepackage{caption}
\usepackage{subcaption}
\usepackage{tabularx}
\usepackage{xcolor}
\usepackage[hidelinks]{hyperref}
\usepackage[nameinlink,noabbrev]{cleveref}
\usepackage{natbib}
\usepackage{placeins}
\usepackage{float}
\usepackage{pgfplots}
\pgfplotsset{compat=1.18}

\title{Instruction-Tuned, but Not More Verifiable Instruction-Following: A Cross-Task Diagnosis for LoRA Adapters}

\author{
  Junyi Zou \\
  Zjydiary Group \\
  \texttt{zoujunyi@zjydiary.cn}
}
\date{}

\begin{document}

\maketitle

\begin{abstract}
Adapters are often selected and deployed based on nominal labels (e.g., \emph{instruction-tuned}), which implicitly suggest what capability improves after adaptation.
We test whether nominal training objectives reliably align with realized cross-task capability gains by evaluating the same LoRA adapter across tasks.
Our strongest evidence is tied to \emph{strict, automatically verifiable instruction following} as measured by IFEval: across multiple seeds, base models, and LoRA settings, nominal labels recurrently—but not universally—fail to predict improvements on this verifiable target, with clear configuration sensitivity including a near-zero or negative case.
As an illustrative strongest-case example in a controlled instruction-versus-numeric setting, an instruction-tuned adapter substantially improves off-target NM-based numeric benchmark performance (0.133$\rightarrow$0.632) while not improving verifiable instruction following on IFEval (ILA: 0.313$\rightarrow$0.271; PLA: 0.250$\rightarrow$0.143; values rounded to three decimals).
We refer to this nominal--realized mismatch pattern as \emph{capability drift} as a descriptive label.
The mismatch is visible in the raw cross-task performance matrix; we use a drift score only as a compact summary in the same units as the underlying metrics, not as a new formal metric contribution.
Evidence from broader instruction-following benchmarks is benchmark-dependent and mixed, reflecting heterogeneity in how instruction following is operationalized; we therefore do not treat cross-benchmark agreement as a premise.
Overall, the practical takeaway is to perform routine cross-task evaluation before deployment and to avoid treating nominal labels as reliable capability proxies.
\end{abstract}

\section{Introduction}\label{sec:introduction}
LoRA adapters \citep{hu2021loralowrankadaptationlarge} are often trained with a nominal task label---\emph{instruction-tuned}, \emph{numeric-reasoning-tuned}, etc.---and then treated as if that label reliably describes what improves after deployment.
In this paper, our strongest evidence is explicitly bounded to \emph{strict, automatically verifiable instruction following} (IFEval) rather than to all notions of instruction following.
We ask a deployment-facing question: do nominal training objectives reliably predict realized cross-task capability gains, especially on verifiable compliance that can be audited automatically?

We frame the work as an empirical cross-task diagnosis: evaluate the same adapter across tasks and compare its nominal-objective gain to its realized gains elsewhere.
Across multiple seeds, base models, and LoRA settings in our runs, we find the nominal--realized mismatch recurrent but not universal and clearly configuration-sensitive, including a near-zero or slightly negative case.
\Cref{tab:main} provides an illustrative strongest-case example: an \emph{instruction-tuned} adapter substantially improves off-target NM-based numeric benchmark performance while not improving strict verifiable instruction following on IFEval \citep{zhou2023instructionfollowingevaluationlargelanguage}.
The primary empirical support, however, comes from the robustness pattern across seeds/models/settings summarized in \Cref{tab:robustness} and Appendix \Cref{tab:robustness_drift}.

Instruction tuning is widely used to improve controllability \citep{zhang2023instructiontuning}, and instruction following can sometimes be elicited via prompting even without instruction tuning \citep{brown2020languagemodelsfewshot}.
But improved helpfulness or instruction-following behavior in broad settings does not imply improved \emph{verifiable} compliance under strict constraints \citep{zhou2023instructionfollowingevaluationlargelanguage,zhang2023instructiontuning}.
We therefore treat benchmark heterogeneity as central rather than incidental: different benchmarks operationalize ``instruction following'' differently, so agreement across them should not be assumed a priori.

We focus on empirical cross-task diagnosis rather than adapter composition, mechanistic attribution, or metric innovation.
We use \emph{capability drift} only as a descriptive label for the observed nominal--realized mismatch pattern.
The practical motivation is that nominal labels are used in model selection and deployment; cross-task evaluation provides a low-cost check for unintended or off-target shifts.

\paragraph{Contributions.}
We make three restrained contributions aligned with our experimental evidence:
\begin{enumerate}
  \item We formulate a deployment-relevant cross-task diagnostic framing: test whether an adapter's nominal label/training objective aligns with realized cross-task capability gains under evaluation.
  \item We provide robustness evidence across seeds, base models, and LoRA settings, showing the mismatch recurrent but configuration-sensitive in magnitude, including a near-zero or slightly negative case.
  \item We give a controlled counterexample: an instruction-tuned adapter can substantially improve off-target NM-based numeric benchmark performance while not improving strict, automatically verifiable instruction following on IFEval.
\end{enumerate}
We additionally report supplementary probing and breakdown analyses, but we treat them as descriptive/exploratory rather than as mechanistic evidence.

\section{Related Work}\label{sec:related}
\paragraph{Adapters, composition, and merge side effects.}
Parameter-efficient adaptation via inserted modules (adapters) \citep{hu2021loralowrankadaptationlarge,houlsby2019parameterefficienttransferlearning} supports modular reuse and composition.
Prior work explores sharing and composing task-specific components \citep{pfeiffer2020adapterfusion,pfeiffer2020adapterhub}, including dynamic composition of LoRA modules \citep{huang2023lorahub}, as well as weight-space mixing and merging of fine-tuned models \citep{wortsman2022modelsoups,ilharco2022editingmodels,yadav2023tiesmerging}.
These lines emphasize interoperability and sometimes analyze when combinations succeed or fail; in contrast, our setting does not involve merging multiple trained adapters or composing multiple tasks at inference time, and we instead diagnose the mismatch between a \emph{single} adapter's nominal objective and its realized cross-task capability gains.

\paragraph{Negative transfer and cross-task trade-offs.}
Multi-task and transfer learning can exhibit cooperation as well as competition, where optimizing for one objective harms performance on another \citep{caruana1997multitasklearning,ruder2017overviewmultitasklearning,standley2019whichtaskstogether}.
Our empirical phenomenon is related but more specific: we diagnose a divergence between the \emph{nominal training objective} (e.g., instruction tuning) and the \emph{most improved measured capability} under cross-task evaluation, even without explicit multi-task training.
We use \emph{capability drift} as shorthand for this mismatch; the contribution is the cross-task diagnostic framing and controlled robustness evidence rather than introducing a new formal object.

\paragraph{Instruction tuning and evaluation mismatch.}
Instruction tuning is broad and evaluated with heterogeneous benchmarks and protocols \citep{ouyang2022traininglanguagemodelstofollowinstructions,zhang2023instructiontuning,wang2023camels}.
Strict, automatically verifiable instruction-following benchmarks such as IFEval \citep{zhou2023instructionfollowingevaluationlargelanguage} capture a narrower notion of compliance than preference-based or open-ended evaluations \citep{zheng2023judgingllmasajudge}.
Recent work further highlights that verifiable instruction-following performance can be benchmark-dependent and may not generalize to unseen constraint families \citep{pyatkin2025generalizing}.
Our main quantitative evidence is therefore anchored on verifiable instruction following, and we treat mixed evidence across additional benchmarks as contextual rather than as generalizing to all notions of instruction following.

\paragraph{Overlap, interference, and localized versus distributed explanations.}
Comparing learned changes across fine-tuning runs is often framed in terms of representation similarity and shared directions \citep{kornblith2019similarity}.
Related to this, weight-space ``task vectors'' \citep{ilharco2022editingmodels} provide a lens for describing learned updates and their interactions.
More broadly, interpretability work on superposition and polysemantic features suggests why changes may be distributed rather than localized to a single unit or module \citep{elhage2022toymodelsofsuperposition}.
Our probing follows this spirit but remains deliberately conservative: we test whether a simple localized account emerges, and interpret negative results as evidence against a single-module explanation rather than as a definitive mechanistic conclusion.

\section{Setup}\label{sec:setup}
\subsection{Base model and LoRA adapters}
We consider a fixed base model and two primary LoRA adapters trained with different nominal objectives:
\emph{reason} (numeric-reasoning-tuned) and \emph{instr} (instruction-tuned).
We also train/evaluate a domain adapter as a secondary setting, but it is not a core pillar of the main claim and is deferred to the appendix.
We do not retrain large models; all analyses reuse existing training and evaluation artifacts.

\subsection{Tasks and metrics}
We evaluate three task families:
\textbf{(i) numeric reasoning benchmark}, \textbf{(ii) verifiable instruction following}, and \textbf{(iii) domain QA} (secondary).

\paragraph{Numeric reasoning benchmark.}
We report numeric match (NM), which measures whether the final numeric answer matches the reference.
Throughout the paper, mentions of ``reasoning'' refer to this benchmark's NM-based operationalization, rather than a claim about broad reasoning ability.

\paragraph{Verifiable instruction following (IFEval).}
IFEval \citep{zhou2023instructionfollowingevaluationlargelanguage} evaluates strict, automatically verifiable instruction constraints.
We report two verifiable metrics:
\textbf{instruction-level accuracy (ILA)} and \textbf{prompt-level accuracy (PLA)}.
ILA captures per-instruction constraint satisfaction (aggregated over instructions), while PLA captures whether the full prompt satisfies all constraints.

\subsection{Drift score}
Let $a$ denote an adapter and let $b$ denote a destination task/metric.
Let $\mathcal{M}_t(\cdot)$ be the scalar metric value for task $t$.
For instruction tuning, we use IFEval prompt-level accuracy (PLA) as the target metric in drift score because it directly reflects strict, end-to-end verifiable compliance; we still report ILA throughout.
We define:
\begin{align}
\mathrm{TargetGain}(a) &= \mathcal{M}_{t(a)}(a) - \mathcal{M}_{t(a)}(\mathrm{base}), \\
\mathrm{OffTargetGain}(a,b) &= \mathcal{M}_b(a) - \mathcal{M}_b(\mathrm{base}), \\
\mathrm{DriftScore}(a \rightarrow b) &= \mathrm{OffTargetGain}(a,b) - \mathrm{TargetGain}(a),
\label{eq:drift_score}
\end{align}
where $t(a)$ is the nominal target task of adapter $a$.
Intuitively, a large positive $\mathrm{DriftScore}(a\rightarrow b)$ indicates that off-target gains exceed target-task gains (or that target-task gains are absent while off-target gains are large).

\paragraph{Metric choice rationale.}
We use a strict, automatically verifiable target metric for instruction tuning (IFEval PLA) because our central question is whether nominal instruction tuning translates into \emph{verifiable compliance} rather than broader helpfulness; PLA is an end-to-end criterion that is sensitive to any violated constraint, while ILA provides a complementary per-constraint view.
For the numeric reasoning benchmark, we use numeric match (NM) because it is simple, task-aligned, and directly comparable across model variants in our setup.
We define drift score as a difference of gains to keep the diagnostic interpretable in the same units as the underlying metrics and to avoid introducing additional normalization choices that would require extra assumptions.
We emphasize that drift score is a \emph{diagnostic summary} of cross-task mismatch in this study, not a universal or uniquely correct capability metric, and the central mismatch is already visible in the raw cross-task performance matrix.

\section{Cross-task mismatch under cross-task evaluation}\label{sec:drift}
\subsection{Reading the cross-task table}
Our primary empirical support is the robustness pattern: across multiple seeds, base models, and LoRA settings, nominal labels can diverge from realized cross-task capability gains in a recurrent but configuration-sensitive way (\Cref{tab:robustness} and Appendix \Cref{tab:robustness_drift}).
\Cref{tab:main} is an illustrative strongest-case example that makes the nominal--realized mismatch immediately visible in raw cross-task metrics.
We refer to this mismatch as \emph{capability drift} as a descriptive label.

\begin{table}[t]
\centering
\small
\caption{\textbf{Illustrative strongest-case cross-task evaluation (main setting).}
We report NM-based numeric benchmark performance (NM) and IFEval strict, automatically verifiable instruction following (ILA/PLA), with values rounded to three decimals.
This example provides a concrete view of the mismatch in raw cross-task metrics; robustness results are summarized in \Cref{tab:robustness} and Appendix \Cref{tab:robustness_drift}.}
\label{tab:main}
\vspace{0.25em}
\begin{tabularx}{\linewidth}{l l c c c X}
\toprule
Model & Nominal task & Numeric NM & IFEval ILA & IFEval PLA & Realized change \\
\midrule
Base & -- & 0.133 & 0.313 & 0.250 & Reference point. \\
Reason adapter & Numeric & 0.309 & 0.271 & 0.179 & Numeric gain aligned with target. \\
Instruction adapter & Instruction & 0.632 & 0.271 & 0.143 & Numeric gain, no IFEval gain. \\
\bottomrule
\end{tabularx}
\end{table}

\subsection{Illustrative strongest-case example}
We emphasize that \Cref{tab:main} is illustrative; the claim is established by robustness results in \Cref{sec:robustness}.
The illustrative mismatch is explicit in \Cref{tab:main}:
\textbf{the instruction-tuned adapter improves off-target NM-based numeric benchmark performance substantially while failing to improve strict, automatically verifiable instruction following on IFEval.}
NM is a benchmark-specific operationalization of numeric performance in our setup (numeric match), and we do not interpret NM gains as a claim about broad reasoning shifts.
Concretely, NM rises from 0.133 (base) to 0.632 (instruction adapter), yet IFEval decreases (ILA: 0.313$\rightarrow$0.271; PLA: 0.250$\rightarrow$0.143; values rounded to three decimals).
By contrast, the numeric-reasoning-tuned adapter achieves a smaller NM gain (0.309) and does not exhibit a comparably large off-target improvement.
Alternative operationalizations (e.g., EM) may change the magnitude of effects, but the nominal--realized mismatch is already visible in the raw cross-task matrix.

\subsection{Quantifying drift with drift score}
\begin{figure}[t]
\centering
\includegraphics[width=0.85\linewidth]{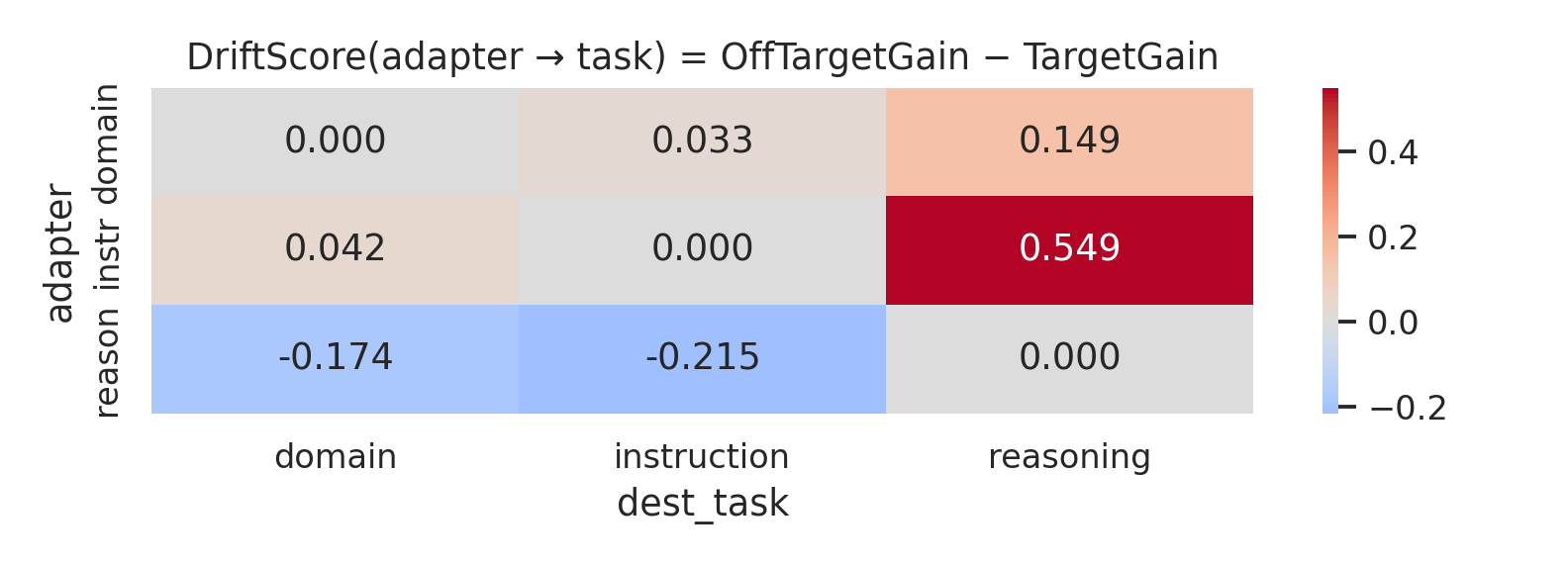}
\caption{\textbf{Drift score summary.}
Heatmap of $\mathrm{DriftScore}(a\rightarrow b)$ computed from cross-task evaluation metrics.
Compact overview of score magnitudes from the same evaluations; see \S\ref{sec:robustness} for context.}
\label{fig:drift}
\end{figure}

\Cref{fig:drift} summarizes drift scores computed from the same evaluation metrics.
Instruction$\rightarrow$numeric reasoning is the most pronounced case: because the instruction adapter's target-task metric (IFEval PLA) does not improve while its off-target NM gain on the numeric reasoning benchmark is large, the drift score is high and positive.
The drift score is a compact summary; the central empirical point does not depend on introducing a new metric beyond the underlying cross-task matrix.

\section{Robustness Across Seeds, Models, and Benchmarks}\label{sec:robustness}
\begin{figure}[t]
\centering
\begin{tikzpicture}
\begin{axis}[
width=0.95\linewidth,
height=0.55\linewidth,
xlabel={TargetGain on IFEval PLA (instr$-$base)},
ylabel={OffTargetGain on numeric NM (instr$-$base)},
xmin=-0.20, xmax=0.02,
ymin=-0.06, ymax=0.62,
grid=both,
legend style={at={(0.02,0.98)},anchor=north west,draw=none,fill=none,font=\small},
]
\addplot[black,dashed,domain=-0.20:0.02,samples=2] {x};
\addplot[black!50,thin] coordinates {(0,-0.06) (0,0.62)};
\addplot[black!50,thin] coordinates {(-0.20,0) (0.02,0)};
\addplot[only marks,mark=*,mark size=1.8pt,color=blue] coordinates {(-0.107,0.094) (-0.107,0.487) (-0.107,0.499) (-0.107,0.516) (-0.071,0.506) (-0.071,0.506) (-0.071,0.517) (-0.071,0.520) (0.000,0.156) (0.000,0.516)};
\addlegendentry{Seed sweep}
\addplot[only marks,mark=triangle*,mark size=2.0pt,color=orange] coordinates {(-0.143,0.344) (-0.107,0.303) (-0.107,0.319) (-0.107,0.500) (-0.107,0.519) (-0.107,0.546) (-0.107,0.565) (-0.107,0.575) (-0.071,0.306) (-0.071,0.343) (-0.071,0.213) (-0.071,0.278) (-0.071,0.516) (-0.071,0.534) (-0.071,0.545) (-0.071,0.546) (-0.036,0.445) (-0.036,0.466) (-0.036,0.300) (0.000,-0.040) (0.000,0.471) (0.000,0.489)};
\addlegendentry{Model/setting sweep}
\addplot[only marks,mark=square*,mark size=2.0pt,color=green!50!black] coordinates {(-0.179,0.306) (-0.107,0.499) (-0.107,0.567)};
\addlegendentry{Benchmark-suite slice}
\end{axis}
\end{tikzpicture}
\caption{\textbf{Robustness quadrant plot (primary evidence).}
Per-run target gain on IFEval PLA (instr$-$base) versus off-target gain on numeric NM (instr$-$base) for the robustness evaluations underlying \Cref{tab:robustness} and Appendix \Cref{tab:robustness_drift}.
Most points lie above $y{=}x$ (positive drift), with occasional near-zero or negative cases.}
\label{fig:robustness-quadrant}
\end{figure}
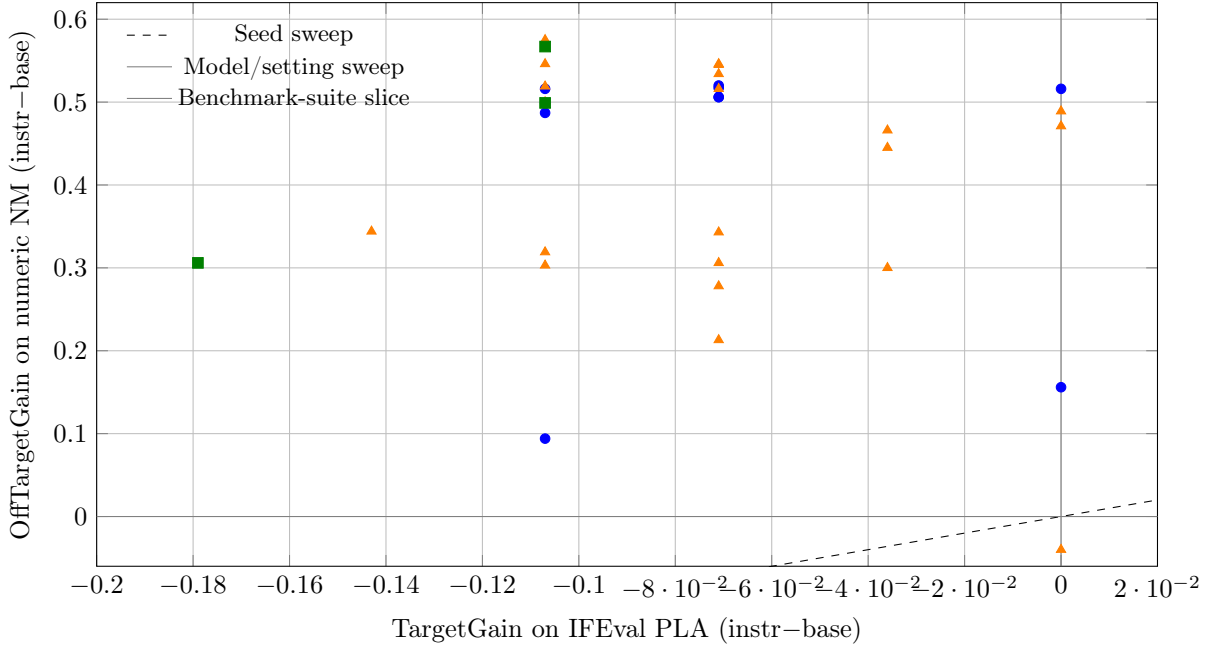

Nominal labels can diverge from realized cross-task gains across seeds, base models, and LoRA settings, in a recurrent but configuration-sensitive way that includes a near-zero or slightly negative case.
Throughout this section, drift score uses the numeric reasoning benchmark NM as the off-target metric and IFEval PLA as the verifiable target metric (\cref{eq:drift_score}).
\Cref{tab:robustness} summarizes the most directly comparable robustness slices.
\Cref{fig:robustness-quadrant} visualizes this primary claim directly using the same robustness evaluations summarized in \Cref{tab:robustness} and Appendix \Cref{tab:robustness_drift}.

\paragraph{Multi-seed robustness.}
On Qwen3-8B, the drift score remains positive across five seeds under two LoRA ranks, with non-trivial but bounded variance (e.g., mean 0.511$\pm$0.178 for $r{=}16$).
This supports that the main mismatch is not a single-seed artifact in this setting.

\paragraph{Multi-model and multi-setting robustness.}
Across multiple base models under a shared setting ($r{=}16$ with attention+MLP modules), drift scores remain positive with small across-seed variance for models where multiple seeds are available.
Across additional LoRA settings, most observed drift scores remain positive, but we also identify configuration sensitivity: one model/setting yields a slightly negative drift score.
Overall, drift magnitude depends on the base model and adapter setting, including a near-zero or slightly negative case.

\paragraph{Additional verifiable instruction-following benchmarks.}
We also run a benchmark suite including FollowBench and IFBench for the same base/adapters.
The results are benchmark-dependent: IFBench and IFEval both emphasize stricter, automatically verifiable compliance and are qualitatively consistent here (no improvement for the instruction adapter), while FollowBench operationalizes a different notion and improves.
We treat this heterogeneity as an insight rather than as a contradiction: cross-benchmark agreement is not a premise because ``instruction following'' is not a single operationalization.
Accordingly, we report the multi-benchmark evidence as contextual and supplementary (\Cref{tab:app-bench}) and keep the strongest claim explicitly tied to strict, automatically verifiable instruction following on IFEval.

\begin{table}[t]
\centering
\small
\caption{\textbf{Robustness summary of drift score (instr$\rightarrow$numeric reasoning).}
Drift score remains positive across five seeds (Qwen3-8B) and across multiple base models and LoRA settings, with one model/setting showing near-zero or slightly negative drift (configuration sensitivity).}
\label{tab:robustness}
\vspace{0.25em}
\input{robustness_summary_table.tex}
\end{table}

\section{Selective Shifts in Verifiable Instruction Following}\label{sec:selective}
\begin{figure}[!htbp]
\centering
\begin{subfigure}[t]{0.48\linewidth}
  \centering
  \includegraphics[width=\linewidth]{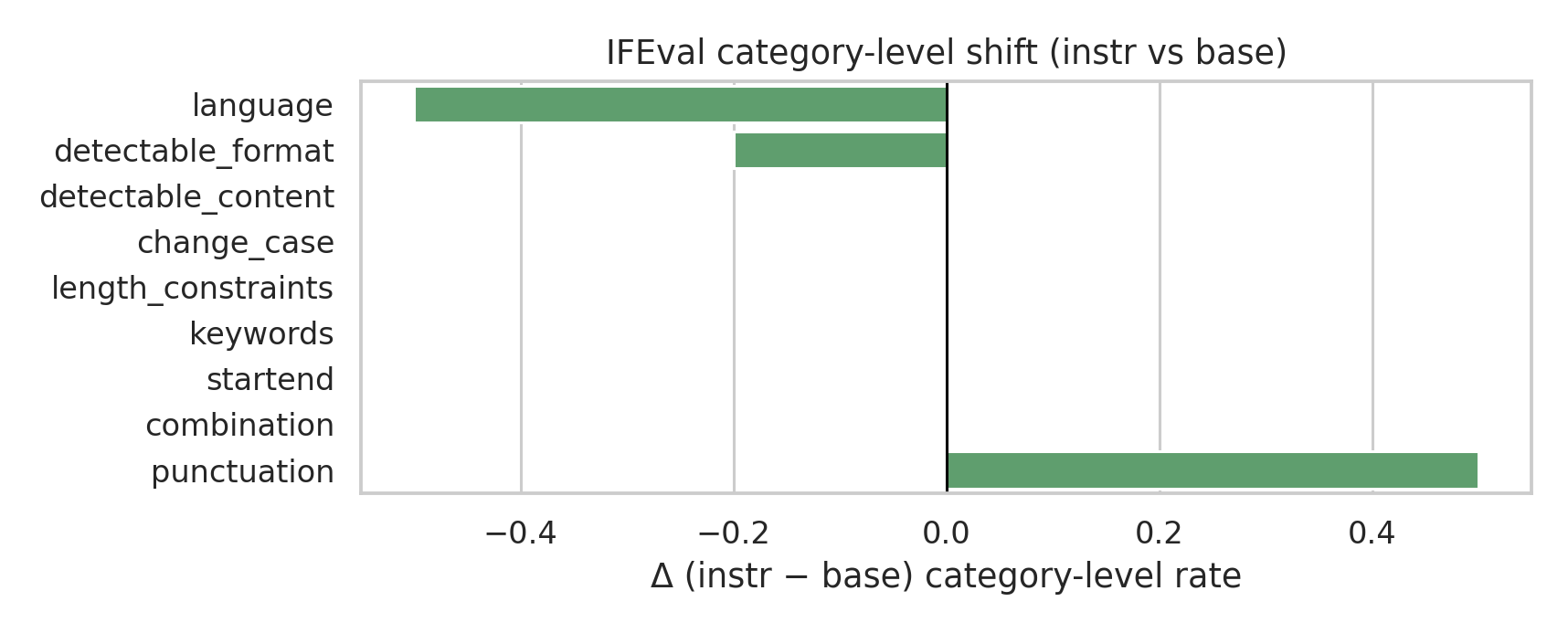}
  \caption{Category-level $\Delta$ (instr$-$base).}
\end{subfigure}\hfill
\begin{subfigure}[t]{0.48\linewidth}
  \centering
  \includegraphics[width=\linewidth]{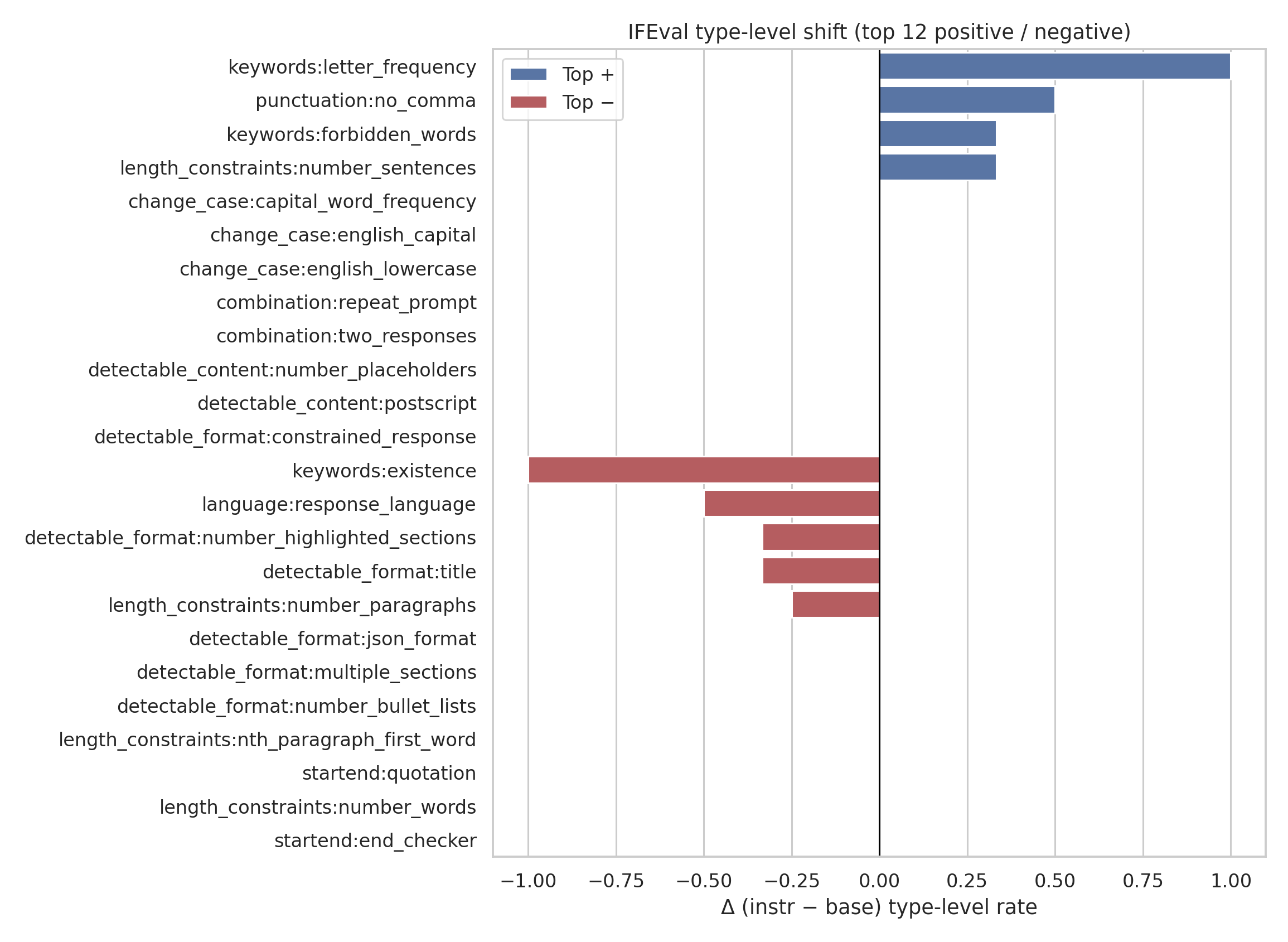}
  \caption{Type-level top positive/negative shifts.}
\end{subfigure}
\caption{\textbf{Selective shifts in verifiable instruction following (IFEval).}
Category-level and type-level shifts for the instruction adapter relative to base on IFEval.
This breakdown is qualitative; low-support types should be interpreted cautiously.}
\label{fig:ifeval}
\end{figure}

The aggregate IFEval metrics (ILA/PLA) hide which verifiable constraints change.
We therefore inspect IFEval category- and type-level breakdowns for the instruction adapter relative to base (\cref{fig:ifeval}).
These breakdowns are descriptive and qualitative: low-support categories or types should be interpreted cautiously, and per-type differences are not intended as high-confidence estimates.

Two representative degradations illustrate the pattern.
At the category level, \textbf{language} constraints drop sharply (rate 0.5$\rightarrow$0.0; $\Delta=-0.5$), and \textbf{detectable\_format} also decreases ($\Delta=-0.2$).
At the type level, \texttt{keywords:existence} shows a large negative shift ($\Delta=-1.0$).
At the same time, some types improve (e.g., \texttt{keywords:letter\_frequency} and \texttt{punctuation:no\_comma} in the top positive shifts).

Overall, \textbf{the aggregate non-improvement on IFEval is not uniformly distributed across verifiable instruction types}; this breakdown is descriptive and does not by itself identify why the mismatch occurs.

\FloatBarrier

\section{Probing Possible Localized Explanations}\label{sec:probing}
This section is supplementary and exploratory.
We probe whether a simple localized account emerges (e.g., a small set of modules or a dominant direction), but we do not treat this as mechanistic or causal evidence.

\paragraph{Summary.}
We observe modest geometric overlap between instruction- and numeric-reasoning-tuned LoRA updates in a small number of upper-layer modules, but the overlap is not uniformly high across modules.

\paragraph{Takeaway.}
Localized functional tests do not provide strong causal evidence linking any particular module/direction to the drift behavior.
Overall, these probes do not support a simple localized explanation in this setup; they should be read as exploratory diagnostics, with additional details in Appendix \Cref{fig:app-geom,fig:app-func}.

\section{Discussion and Limitations}\label{sec:discussion}
\paragraph{Nominal labels can diverge from realized gains.}
In our evaluated settings, an adapter's nominal label or nominal training objective does not reliably predict realized cross-task capability gains under deployment-relevant evaluation.
Cross-task evaluation can reveal nominal--realized mismatches that are invisible when only the nominal target task is reported.

\paragraph{Instruction tuning vs verifiable instruction following.}
Instruction tuning should not be conflated with improved strict, automatically verifiable instruction following.
In our illustrative strongest-case example, off-target NM-based numeric benchmark performance improves substantially while IFEval does not.
IFEval targets strict, automatically verifiable compliance and is narrower than broader notions of helpful or conversational instruction following; benchmark-dependent operationalizations therefore matter when interpreting whether a nominal--realized mismatch is present.

\paragraph{Plausible hypotheses (speculative).}
One possibility is that instruction tuning shifts response priors (e.g., completion style or answer-format regularity) in ways that help a benchmark-specific numeric task while leaving strict constraint satisfaction unchanged.
We do not adjudicate such hypotheses here, and we avoid interpreting the results as evidence about broad reasoning shifts.

\paragraph{Limitations.}
First (empirical scope), our findings are empirical and specific to the models, tasks, and adapters evaluated here.
Second (claim scope), our strongest claim is tied to strict, automatically verifiable instruction following (IFEval) rather than to broader notions of instruction following.
Third (benchmark heterogeneity), broader instruction-following benchmarks operationalize different targets and yield benchmark-dependent evidence in our suite, so cross-benchmark agreement should not be assumed.
Fourth (probing inconclusive), our exploratory probing remains non-causal and inconclusive: it does not support a simple localized explanation, but it does not establish a mechanism.
Fifth (metric choice), we operationalize benchmark-specific numeric performance via NM and summarize mismatch with drift score; alternative operationalizations (e.g., EM or different instruction-following aggregations) can change magnitudes, as shown in Appendix \Cref{tab:app-domain,tab:metric-sensitivity}.
We also observe configuration sensitivity, including a near-zero or slightly negative case, which should be interpreted as evidence that the pattern is recurrent but not universal.

\section{Conclusion}\label{sec:conclusion}
We present an empirical cross-task diagnosis for LoRA adapters.
Across multiple seeds, base models, and LoRA settings, we observe a recurrent but not universal nominal--realized mismatch, with the strongest evidence tied to strict, automatically verifiable instruction following (IFEval).
An illustrative case shows large off-target NM-based numeric gains without IFEval improvement.
The practical takeaway is to perform routine cross-task evaluation before deployment and avoid treating nominal labels as reliable capability proxies.

\bibliographystyle{plainnat}
\bibliography{references}

\clearpage
\appendix
\section{Secondary results and additional figures}\label{app:secondary}
\subsection{Domain adapter as a secondary setting}
We evaluate a domain-tuned adapter as an additional setting, but do not use it as core evidence for the main claim.
\Cref{tab:app-domain} provides an extended cross-task view including the domain adapter and additional metrics.

\begin{table}[H]
\centering
\small
\caption{\textbf{Extended cross-task evaluation including a domain adapter (secondary).}
This table is provided for completeness and is not central to the paper's primary robustness claim, which is established by \Cref{tab:robustness} and Appendix \Cref{tab:robustness_drift}; \Cref{tab:main} is illustrative. All values are reported as fractions in $[0,1]$; numeric benchmark performance is reported using both exact match (EM) and numeric match (NM).}
\label{tab:app-domain}
\vspace{0.5em}
\input{main_cross_task_table.tex}
\end{table}

\subsection{Additional benchmarks: FollowBench and IFBench}
To test whether the instruction adapter improves instruction-following beyond IFEval, we report two additional benchmarks from a benchmark suite: FollowBench (rule-based score summary) and IFBench (strict prompt-level accuracy).
Across three base models under the same setting, IFBench aligns more closely with IFEval because both emphasize stricter, automatically verifiable compliance, and we observe no improvement for the instruction adapter on these strict criteria.
By contrast, FollowBench operationalizes a different notion and improves.
We report these results as contextual and supplementary: benchmark-dependent operationalizations matter, so cross-benchmark agreement is not assumed and does not expand the paper's strongest claim beyond IFEval.

\subsection{Metric sensitivity: target metric PLA vs ILA}
Our main diagnostic uses IFEval PLA as the target metric for instruction tuning because it is an end-to-end strict criterion, and we report ILA throughout as a complementary view.
To illustrate sensitivity to this choice without introducing new experiments, we recompute the instruction$\rightarrow$numeric reasoning drift score in the main setting using ILA in place of PLA (all values derived from \Cref{tab:main}).
The mismatch direction is unchanged: off-target NM gains on the numeric reasoning benchmark remain large while verifiable instruction-following target gains remain absent, and the drift score remains strongly positive; only the magnitude varies with the target aggregation.

\begin{table}[H]
\centering
\small
\caption{\textbf{Target-metric sensitivity check (PLA vs ILA).}
Instruction$\rightarrow$numeric reasoning drift score recomputed from \Cref{tab:main} using IFEval PLA or ILA as the instruction target metric.}
\label{tab:metric-sensitivity}
\vspace{0.25em}
\begin{tabular}{lccc}
\toprule
Target metric & OffTargetGain (NM) & TargetGain (IFEval) & DriftScore (instr$\rightarrow$numeric reasoning) \\
\midrule
PLA & 0.499 & $-0.107$ & 0.606 \\
ILA & 0.499 & $-0.042$ & 0.541 \\
\bottomrule
\end{tabular}
\end{table}

\begin{table}[H]
\centering
\small
\caption{\textbf{Supplementary multi-benchmark instruction following.} FollowBench rule-based mean and IFBench strict PLA for base vs instruction adapter. Values are computed from benchmark-suite outputs. FollowBench, IFBench, and IFEval differ in scale and emphasis, so absolute values should not be compared across benchmarks; in particular, FollowBench rule mean should be interpreted according to its scoring definition.}
\label{tab:app-bench}
\vspace{0.25em}
\begin{tabular}{lcccc}
\toprule
Model & \shortstack{IFBench\\PLA (strict)\\base} & \shortstack{IFBench\\PLA (strict)\\instr} & \shortstack{FollowBench\\rule mean\\base} & \shortstack{FollowBench\\rule mean\\instr} \\
\midrule
Qwen3-8B & 0.122 & 0.112 & 0.000 & 4.000 \\
Qwen3-14B & 0.143 & 0.116 & 0.000 & 2.000 \\
Llama-3.1-8B-Instruct & 0.139 & 0.119 & 4.000 & 4.500 \\
\bottomrule
\end{tabular}
\end{table}

\subsection{Probing figures}
These figures are placed in the appendix to keep the main paper focused on the empirical diagnosis and robustness results.
They provide supporting context for \Cref{sec:probing}: how to read them is as \emph{diagnostics} for whether a simple localized account might emerge.
Consistent with the main takeaway, the patterns suggest at most modest geometric overlap in a few upper-layer modules, while the functional interventions do not yield strong localized causal evidence; we therefore avoid interpreting these plots as identifying a single-module mechanism for drift.

\begin{figure}[t]
\centering
\includegraphics[width=0.85\linewidth]{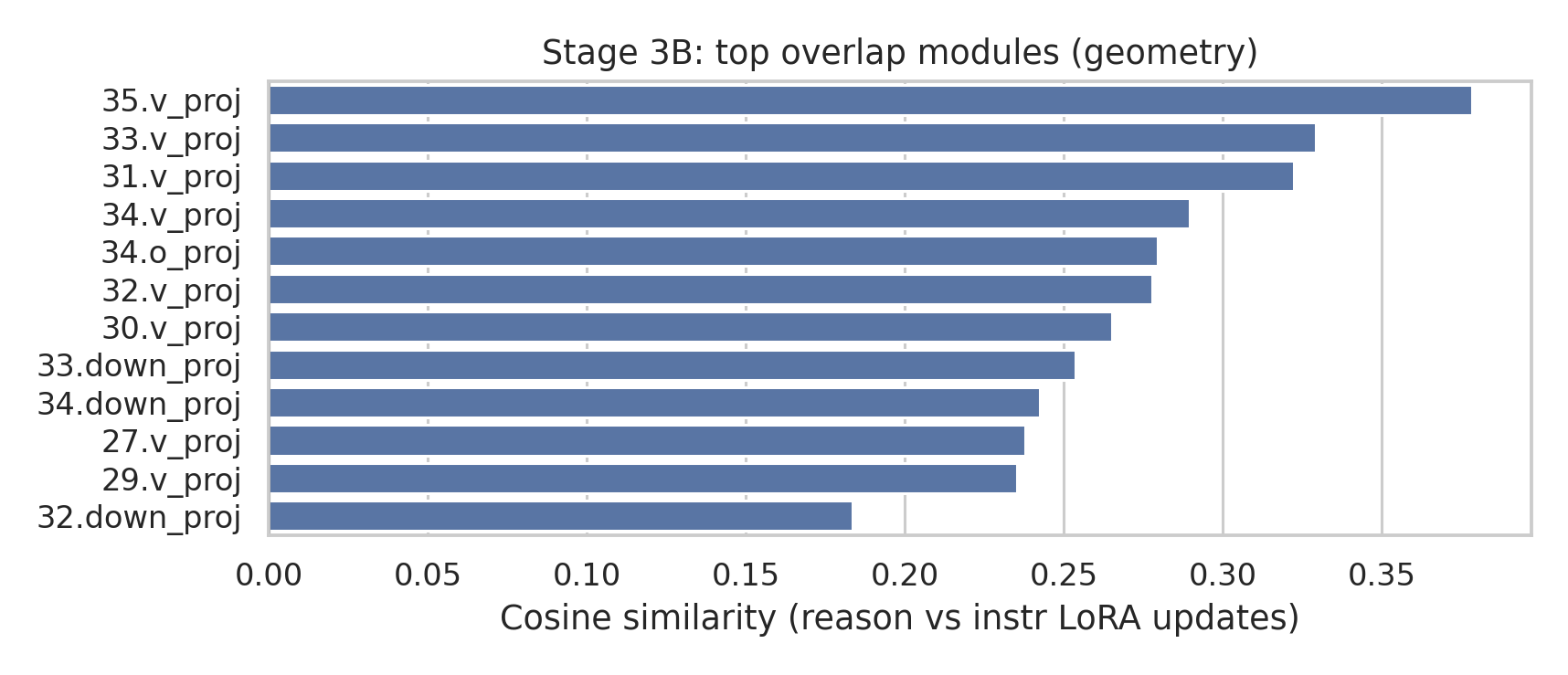}
\caption{\textbf{Upper-layer geometry diagnostics.}
Top modules by geometric similarity between instruction- and numeric-reasoning-tuned LoRA updates.
This figure provides preliminary evidence but does not establish a localized causal mechanism.}
\label{fig:app-geom}
\end{figure}

\begin{figure}[t]
\centering
\includegraphics[width=0.85\linewidth]{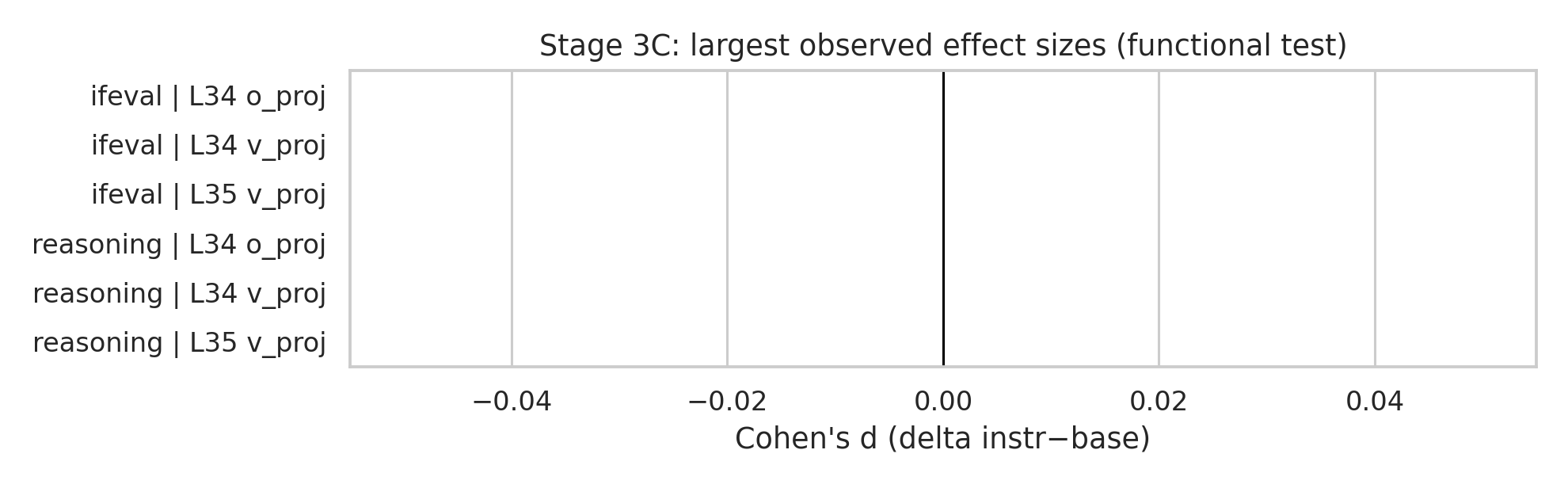}
\caption{\textbf{Functional probing summary with null or small effects.}
Largest observed effect sizes from localized functional probing.
We do not find strong localized causal evidence that explains the mismatch pattern in this setup.}
\label{fig:app-func}
\end{figure}

\subsection{Robustness tables}\label{app:robustness}
This subsection provides the full robustness numbers referenced in \Cref{sec:robustness}, including per-seed and per-setting slices.
They are intended as supporting material for transparency and reproducibility, not as additional main claims beyond the summaries already reported in the main text.

\input{robustness_drift_summary.tex}
\paragraph{Benchmark-suite run status (bookkeeping).}
All benchmark-suite evaluations referenced in the paper completed successfully for the included model/seed/setting combinations; we omit the full per-run status table here because it is bookkeeping rather than a scientific result.

\end{document}

%% file: robustness_summary_table.tex
\begin{tabular}{l l l c c}
\toprule
Slice & Model & LoRA setting & $n$ & Drift score ($\uparrow$) \\
\midrule
Seeds & qwen3\_8b & r16\_attnmlp\_do005\_lrmain & 5 & 0.511$\pm$0.178 \\
Seeds & qwen3\_8b & r8\_attnmlp\_do005\_lrmain & 5 & 0.495$\pm$0.150 \\
\midrule
Models & qwen3\_14b & r16\_attnmlp\_do005\_lrmain & 2 & 0.667$\pm$0.014 \\
Models & qwen3\_8b & r16\_attnmlp\_do005\_lrmain & 2 & 0.597$\pm$0.010 \\
Models & llama31\_8b\_instruct & r16\_attnmlp\_do005\_lrmain & 2 & 0.394$\pm$0.016 \\
Models & qwen25\_7b\_instruct & r16\_attnonly\_do0\_lrsmall & 1 & -0.040 \\
\bottomrule
\end{tabular}

%% file: main_cross_task_table.tex
\setlength{\tabcolsep}{4pt}
\begin{tabular}{lcccccc}
\toprule
 & \shortstack{Numeric\\EM} & \shortstack{Numeric\\NM} & \shortstack{IFEval\\ILA} & \shortstack{IFEval\\PLA} & \shortstack{Domain\\Acc} & \shortstack{Domain\\MC} \\
\midrule
base & 0.000 & 0.133 & 0.312 & 0.250 & 0.000 & 0.000 \\
reason & 0.042 & 0.306 & 0.271 & 0.179 & 0.000 & 0.000 \\
instr & 0.000 & 0.640 & 0.271 & 0.143 & 0.000 & 0.000 \\
domain & 0.000 & 0.281 & 0.345 & 0.148 & 0.000 & 0.000 \\
\bottomrule
\end{tabular}

%% file: robustness_drift_summary.tex
\begin{table}[t]
\centering
\small
\caption{\textbf{Robustness: drift score summary (instr$\rightarrow$numeric reasoning).} Drift score mean and standard deviation across seeds (when $n>1$); values are rounded to three decimals.}
\label{tab:robustness_drift}
\begin{tabular}{lllrc}
\toprule
Stage & Model & Setting & $n$ & Drift score (mean $\pm$ std) \\
\midrule
benchmark\_suite & llama31\_8b\_instruct & r16\_attnmlp\_do005\_lrmain & 1 & $0.485$ \\
benchmark\_suite & qwen3\_14b & r16\_attnmlp\_do005\_lrmain & 1 & $0.674$ \\
benchmark\_suite & qwen3\_8b & r16\_attnmlp\_do005\_lrmain & 1 & $0.606$ \\
model\_sweep & deepseek\_r1\_distill\_qwen\_7b & r16\_attnonly\_do0\_lrsmall & 1 & $0.284$ \\
model\_sweep & deepseek\_r1\_distill\_qwen\_7b & r32\_attnmlp\_do005\_lrmain & 1 & $0.350$ \\
model\_sweep & llama31\_8b\_instruct & r16\_attnmlp\_do005\_lrmain & 2 & $0.394 \pm 0.016$ \\
model\_sweep & llama31\_8b\_instruct & r16\_attnonly\_do0\_lrsmall & 1 & $0.426$ \\
model\_sweep & llama31\_8b\_instruct & r32\_attnmlp\_do005\_lrmain & 1 & $0.336$ \\
model\_sweep & llama31\_8b\_instruct & r8\_attnonly\_do0\_lrmain & 2 & $0.451 \pm 0.036$ \\
model\_sweep & qwen25\_7b\_instruct & r16\_attnonly\_do0\_lrsmall & 1 & $-0.040$ \\
model\_sweep & qwen25\_7b\_instruct & r32\_attnmlp\_do005\_lrmain & 1 & $0.471$ \\
model\_sweep & qwen3\_14b & r16\_attnmlp\_do005\_lrmain & 2 & $0.667 \pm 0.014$ \\
model\_sweep & qwen3\_14b & r16\_attnonly\_do0\_lrsmall & 1 & $0.617$ \\
model\_sweep & qwen3\_14b & r32\_attnmlp\_do005\_lrmain & 1 & $0.672$ \\
model\_sweep & qwen3\_14b & r8\_attnonly\_do0\_lrmain & 2 & $0.611 \pm 0.006$ \\
model\_sweep & qwen3\_8b & r16\_attnmlp\_do005\_lrmain & 2 & $0.597 \pm 0.010$ \\
model\_sweep & qwen3\_8b & r16\_attnonly\_do0\_lrsmall & 1 & $0.481$ \\
model\_sweep & qwen3\_8b & r32\_attnmlp\_do005\_lrmain & 1 & $0.626$ \\
model\_sweep & qwen3\_8b & r8\_attnonly\_do0\_lrmain & 2 & $0.495 \pm 0.006$ \\
seed\_sweep & qwen3\_8b & r16\_attnmlp\_do005\_lrmain & 5 & $0.511 \pm 0.178$ \\
seed\_sweep & qwen3\_8b & r8\_attnmlp\_do005\_lrmain & 5 & $0.495 \pm 0.150$ \\
\bottomrule
\end{tabular}
\end{table}